\documentclass[lettersize,journal]{IEEEtran}
\usepackage{amsmath,amsfonts}
\usepackage{algorithm}
\usepackage{array}
\usepackage[caption=false,font=normalsize,labelfont=sf,textfont=sf]{subfig}
\usepackage{textcomp}
\usepackage{stfloats}
\usepackage{url}
\usepackage{verbatim}
\usepackage{graphicx}
\usepackage{cite}
\usepackage{xcolor} 
\usepackage{booktabs}
\usepackage{multirow}
\usepackage[switch]{lineno}
\usepackage[utf8]{inputenc}
\usepackage{times}
\usepackage{soul}
\usepackage{algpseudocode}
\usepackage{orcidlink}

\begin{document}

\title{Square Superpixel Generation and Representation Learning via Granular Ball Computing}

\author{%
    Shuyin~Xia\orcidlink{0000-0001-5993-9563},~\IEEEmembership{Senior~Member,~IEEE},~
    Meng~Yang\orcidlink{0009-0005-8311-092X},~
    Dawei~Dai*\orcidlink{0000-0002-8431-4431},~
    Fan~Chen\orcidlink{0009-0007-2163-5557},~
    Shilin~Zhao,~
    
    Junwei~Han,~\IEEEmembership{Fellow,~IEEE},~
    Xinbo~Gao\orcidlink{0000-0002-7985-0037},~\IEEEmembership{Fellow,~IEEE},~
    Guoyin~Wang\orcidlink{0000-0002-8521-5232},~\IEEEmembership{Senior~Member,~IEEE},
    and Wen~Lu%

    \thanks{This work was supported by Chongqing Key Laboratory of Computational Intelligence, 
    Key Laboratory of Cyberspace Big Data Intelligent Security, Ministry of Education, 
    Sichuan-Chongqing Co-construction Key Laboratory of Digital Economy Intelligence, 
    and Key Laboratory of Big Data Intelligent Computing, 
    Chongqing University of Posts and Telecommunications, Chongqing, China.}%
    \thanks{Shuyin Xia, Meng Yang, Dawei Dai, Fan Chen, Shilin Zhao and Junwei Han are with 
    Chongqing University of Posts and Telecommunications, Chongqing, China.}%
    \thanks{Xinbo Gao and Wen Lu are with Xidian University, Xi'an, China.}%
    \thanks{Guoyin Wang is with Chongqing Normal University, Chongqing, China.}%
    \thanks{*Corresponding author: Dawei Dai (dw\_dai@163.com).}%
    \thanks{This article is an extended version of our conference paper published in CISAT 2025\cite{zhao2025square}.}%
}

% <-this % stops a space

\markboth{Journal of \LaTeX\ Class Files,~Vol.~14, No.~8, August~2021}%
{Shell \MakeLowercase{\textit{et al.}}: A Sample Article Using IEEEtran.cls for IEEE Journals}


\maketitle

\begin{abstract}
Superpixels provide a compact region-based representation that preserves object boundaries and local structures, and have therefore been widely used in a variety of vision tasks to reduce computational cost. However, most existing superpixel algorithms produce irregularly shaped regions, which are not well aligned with regular operators such as convolutions. Consequently, superpixels are often treated as an offline preprocessing step, limiting parallel implementation and hindering end-to-end optimization within deep learning pipelines. Motivated by the adaptive representation and coverage property of granular-ball computing, we develop a square superpixel generation approach. Specifically, we approximate superpixels using multi-scale square blocks to avoid the computational and implementation difficulties induced by irregular shapes, enabling efficient parallel processing and learnable feature extraction. For each block, a purity score is computed based on pixel-intensity similarity, and high-quality blocks are selected accordingly. The resulting square superpixels can be readily integrated as graph nodes in graph neural networks  (GNNs) or as tokens in Vision Transformers (ViTs), facilitating multi-scale information aggregation and structured visual representation. Experimental results on downstream tasks demonstrate consistent performance improvements, validating the effectiveness of the proposed method.
\end{abstract}

\begin{IEEEkeywords}
Superpixels, Granular ball computing, Square superpixels, Multi-scale tokenization, Vision Transformers.
\end{IEEEkeywords}

\section{Introduction}

% a parallelizable superpixel generation mechanism with
% learnable features, which can be further embedded into neural
% networks. The comparison between our approach and SLIC
% in image superpixel generation is illustrated in Figure 1.
% In contrast, our method more effectively expresses images,
% amplifies the model’s sensitivity to fine-grained features, and
% can be seamlessly integrated with convolutional networks. The
% main contributions of this work can be summarized as follows.

\begin{figure*}[htbp]
\centering
\includegraphics[width=\textwidth,keepaspectratio]{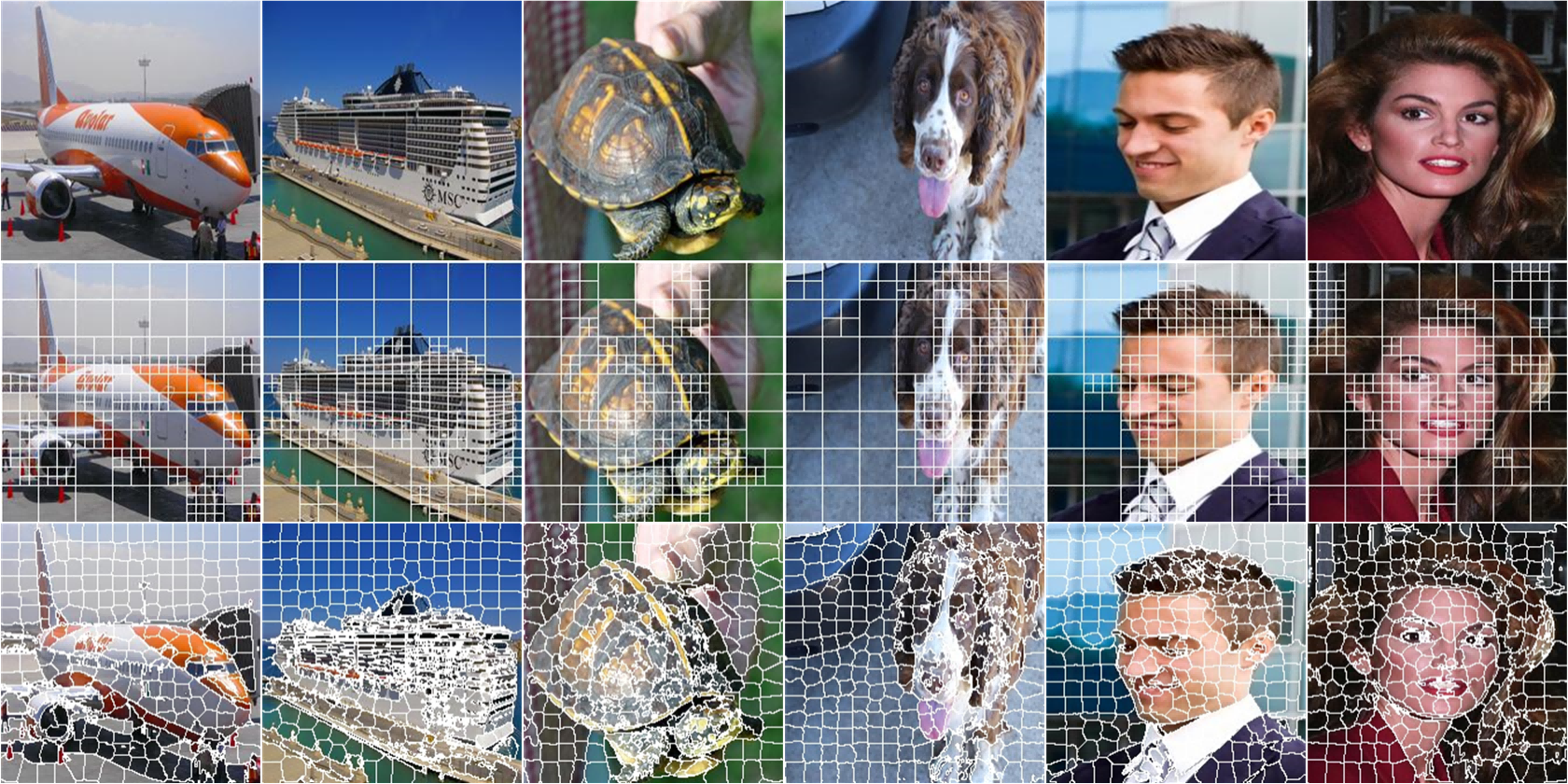} 
\caption{In contrast with SLIC, the first line depicts the input image, the second showcases the superpixel blocks generated by our method, and the last line displays those generated by the SLIC method.}
\label{fig}
\end{figure*}

\IEEEPARstart
Recent years have witnessed remarkable progress in computer vision \cite{cui2023focal, li2025u, cui2025adair}; nevertheless, practical systems in complex scenes still often suffer from limited robustness and interpretability, together with nontrivial computational overhead \cite{vaswani2017attention, dunkel2025cns, haurum2023tokens}. To alleviate the redundancy of pixel-level modeling, superpixels have been widely adopted as a classical region-based representation: by grouping pixels with similar attributes (e.g., color, texture, or intensity) into compact regions, superpixels provide a more efficient mid-level abstraction that preserves boundaries and local structures while offering useful structural priors for downstream tasks.

Owing to these advantages, superpixels have been extensively used in image segmentation, object tracking, object detection, recognition, and classification, and are often combined with Graph Neural Networks (GNNs) or Transformers as structured units to explicitly model inter-region relations and local interactions. In particular, for detection scenarios, region-level representations help aggregate local evidence under multi-scale targets and complex backgrounds, highlight object contours, and provide a more compact set of candidate modeling units when computational budgets are constrained.
However, most existing superpixel algorithms produce fragmented and irregular region units. While such irregular shapes can be beneficial for boundary adherence, they also introduce implementation and training costs that are misaligned with modern deep learning pipelines. On the one hand, many methods (e.g., SLIC~\cite{achanta2012slic}) are non-differentiable, making them difficult to incorporate into deep networks for end-to-end optimization. On the other hand, pursuing high boundary accuracy often incurs substantial computational overhead. Moreover, irregular regions are poorly aligned with regular operators such as convolutions, which limits batch-parallel execution and hampers learnable feature modeling. To improve differentiability and task adaptivity, learning-based superpixel methods (e.g., differentiable clustering or FCN-based approaches~\cite{yang2020superpixel}) have been proposed in recent years. However, these methods typically require predicting soft pixel-to-superpixel associations and assignment weights, and they are commonly trained with reconstruction, compactness, or boundary-aware losses. In addition, some approaches impose explicit connectivity constraints or introduce extra steps to enforce connected regions, which further increases implementation and optimization complexity as well as computational and memory costs. Consequently, in many practical settings, superpixel generation and subsequent modeling remain separated into an offline preprocessing stage and an independent learning stage, limiting the potential benefits of superpixels within end-to-end deep frameworks.
Beyond unimodal vision tasks, image–text alignment in multimodal learning poses additional challenges. Images are commonly tokenized as pixels or regular patches, whereas text is represented by discrete semantic symbols. Without structural priors, visual tokens often struggle to establish stable correspondences with textual units, thereby affecting the efficiency and interpretability of joint multimodal representations. This motivates the development of a visual discretization that is both structurally expressive and compatible with the computational form of modern neural networks.

To address these issues, inspired by the adaptive representation and coverage principles of Granular Ball Computing (GBC), we propose a rule-based, non-iterative hierarchical partition-and-selection scheme for data-space partitioning, and further develop a square-superpixel generation mechanism that integrates efficiently with convolutional neural networks. Our approach approximates superpixels with multi-scale square blocks and computes a purity score in the pixel space via an intra-block consistency measure: a statistic extracted from the block center serves as a reference, and pixel similarity to this reference is used to estimate purity and select high-quality regions. We then reuse the feature extraction capability of a backbone network (e.g., ResNet) to produce multi-scale feature representations and select structured tokens according to scale-specific masks, enabling tensor-friendly parallelism and end-to-end integration. The proposed design requires neither pixel-level superpixel annotations nor training an independent superpixel generator. Moreover, the resulting square superpixel blocks can be naturally embedded into GNNs or ViTs to support multi-scale information extraction and structured visual representation.

In contrast to learning-based superpixel methods such as SSN, which typically rely on soft pixel–superpixel associations (assignment) and perform feature aggregation via differentiable clustering or association matrices, our method uses regular blocks as the basic units for block-wise scoring and selection, without explicitly maintaining large-scale soft association matrices. As a result, it is more amenable to tensorized implementation and naturally supports parallel computation and token pruning. Leveraging this property, we embed the structured visual tokens produced by square superpixels into the RT-DETR \cite{lv2023detrs} detection framework  and discard redundant tokens before self-attention to reduce computational cost. Experimental results show that detection performance changes only marginally while the number of tokens is reduced, indicating that the proposed structured token representation preserves key object information while effectively suppressing redundant background content, thereby enabling efficient end-to-end detection.

\begin{itemize}
\item We propose an end-to-end compatible, rule-driven square-superpixel tokenization mechanism that generates structured tokens on-the-fly via a non-iterative, block-wise scoring-and-selection procedure. It avoids explicit soft association matrices, is GPU-parallel friendly, and can be seamlessly integrated into standard deep networks without any extra learnable generator or pre-processing stage.
\item We introduce a multi-granularity visual discretization strategy that unifies square superpixels at different scales into discrete representations directly consumable by graph networks and Transformers, and enables token pruning in RT-DETR to reduce attention computation.
\item We conduct systematic experiments across multiple datasets and diverse network architectures to demonstrate the effectiveness and generalization of the proposed approach, and to validate its efficiency benefits under token pruning settings.
\end{itemize}

\section{Related Work}
\subsection{Superpixels}
Superpixel generation methods can be broadly grouped into non-end-to-end traditional pipelines and end-to-end trainable models. Traditional approaches mainly fall into three representative families. (i) Graph partitioning and energy optimization methods, e.g., Ncuts~\cite{shi2000normalized} and ERS~\cite{liu2011entropy}, produce boundary-aligned over-segmentations by optimizing cut criteria or energy-based objectives. (ii) Fast local clustering and iterative refinement methods, such as SLIC~\cite{achanta2012slic} and SEEDS~\cite{van2012seeds}, generate compact and roughly size-uniform superpixels via efficient local clustering or update rules, offering a favorable efficiency–quality trade-off. (iii) Watershed and region evolution methods, including Watershed~\cite{vincent1991watersheds} and TurboPixels~\cite{levinshtein2009turbopixels}, rely on watershed transforms or region-evolution processes to form boundary-aware region partitions.

In recent years, end-to-end trainable methods have attracted increasing attention. Representative examples include SSN~\cite{jampani2018superpixel} and SpixelFCN~\cite{yang2020superpixel}, which learn pixel-to-superpixel assignments in a differentiable manner and attempt to jointly optimize superpixel generation with downstream tasks. However, end-to-end methods still share several limitations. Soft pixel--superpixel assignment together with feature aggregation increases compute and memory cost at high resolutions, while satisfying geometric constraints such as connectivity and compactness typically requires additional regularization or post-processing. Moreover, proxy objectives such as reconstruction, boundary alignment, and compactness do not always match downstream goals, which makes training more sensitive to hyperparameters and domain shifts.

\begin{figure*}[htbp]
\centering
\includegraphics[width=\textwidth,keepaspectratio]{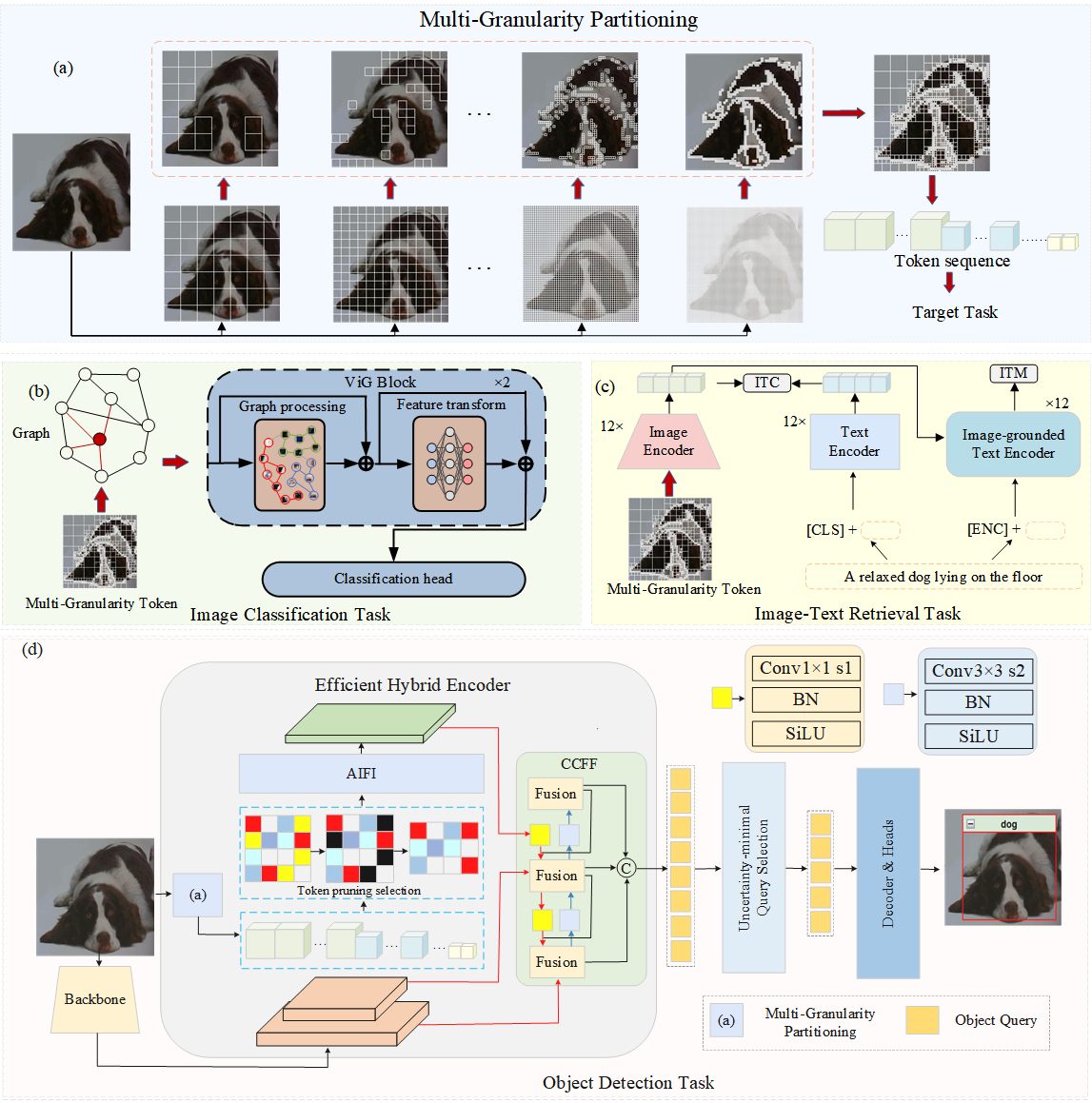} 
\caption{Overview of our superpixel generation approach: (a) Module for multi-scale superpixels; (b) Integration of the module into ViG for image classification; (c) Integration of the module into ViT for image-text retrieval; (d) Integration of the module into ViT for object detection.}
\label{fig}
\end{figure*}

\subsection{GNN for image graph classification}

Most superpixel-based image classification methods currently employ GNNs due to their strengths in modeling irregular structures and capturing local relationships. Existing studies primarily rely on superpixels generated by the SLIC algorithm for posture representation, often using 2D coordinates \cite{AchantaSSLFS12}, linear projections \cite{youn2021dynamic}, or trainable positional embeddings\cite{bae2022superpixel}. However, the superpixels produced by SLIC are overly uniform in shape, lacking sufficient discriminative information.  Shape features can better reflect the representative content of an image, and incorporating shape information has the potential to improve classification performance. The ShapeGNN method proposed by Cosma et al. \cite{cosma2023geometric} combines shape, appearance, and posture features of superpixels, designing an end-to-end graph network. However, its classification performance is limited on low-resolution datasets, such as CIFAR-10, due to insufficient shape information.  

In contrast, our method introduces a purity and adaptive segmentation mechanism, dynamically adjusting the block size of the image, selecting the most representative regions, and filtering out noise. This approach enhances classification accuracy while supporting end-to-end training.

\subsection{Image-Text Retrieval}

Current image-text retrieval tasks primarily rely on contrastive learning to train on large-scale datasets, maximizing the similarity between images and text, as seen in models like CLIP \cite{radford2021learning} and BLIP \cite{li2022blip}. These methods enhance the matching of image-text pairs by learning from noisy large-scale datasets. FLIP \cite{dai202415m}, on the other hand, focuses on data cleaning within the LAION-Face dataset, constructing a large-scale facial dataset and training a model with strong facial recognition capabilities.  

However, these models predominantly perform feature extraction at a coarse-grained level, failing to effectively capture fine-grained image features. To address this issue, FILIP \cite{yao2021filip} introduces an interaction mechanism to optimize image-text pair similarity, while SPARC \cite{bica2024improving} further improves fine-grained alignment through fine-grained contrastive loss. Nevertheless, these approaches still suffer from insufficient precision in fine-grained feature capture and lack interpretability.  

\subsection{Object detection}
Object detection has evolved from region-proposal-based two-stage pipelines to efficient one-stage detectors, and has continued to advance with the development of deep learning. Recently, Vision Transformers have introduced an end-to-end set prediction paradigm for detection. DETR \cite{carion2020end} is a representative framework that removes hand-crafted components such as NMS, but it often suffers from slow convergence and high computational cost. Subsequent variants, such as Deformable DETR \cite{zhu2020deformable} and RT-DETR \cite{lv2023detrs}, improve training efficiency and real-time performance by enhancing attention mechanisms and adopting hybrid encoder designs.

Nevertheless, the main bottleneck of transformer-based detectors still stems from the quadratic complexity of self-attention. To reduce computation, prior work has explored token pruning and token merging to shorten the sequence length, yet these strategies typically lack explicit spatial-structure constraints and interpretability. In contrast, we introduce a purity-driven hierarchical token selection mechanism grounded in  granular ball computing: by explicitly measuring region homogeneity and selecting informative tokens in a coarse-to-fine manner, our approach reduces computation while preserving the spatial structures and object-relevant cues that are crucial for detection.

\subsection{Granular ball computing}

Chen et al. \cite{chen1982topological} pointed out in their research published in Science that human cognition follows the principle of ``global precedence". Building on the theoretical foundation of traditional granular computing, Wang et al.  \cite{wang2017dgcc} pioneered the concept of multigranular cognitive computing, integrating this cognitive principle. Based on multigranular cognitive computing, Xia et al. \cite{xia2020ball} proposed granular-ball computing. This method utilizes ``granular balls" of varying sizes to adaptively cover and represent the sample space, enabling computing and learning based on granular balls instead of points. Granular ball computing has been applied across various artificial intelligence domains,  examples include granular ball reinforcement learning \cite{liu2024unlock}, granular ball classifiers \cite{xia2022efficient,xia2025graph}, granular ball clustering \cite{xia2020ball,xie2024efficient, xie2024mgnr}, and granular ball support vector machines \cite{xia2024gbsvm,quadir2024granular}, and granular ball graph representation \cite{10996538}. This study draws on the concept of granular-balls computing and proposes an innovative superpixel generation method that improves the accuracy of image feature extraction through purity and an adaptive partitioning mechanism. This method dynamically adjusts block sizes, selects representative regions, and filters out noise, thereby enhancing classification performance.

\section{Method}
To facilitate seamless integration of superpixel representations into modern deep learning frameworks while fully exploiting GPU parallelism, we aim to construct a \emph{fixed-budget} structured visual unit that is consistent with the computational form of mainstream operators. In this paper, a square superpixel refers to a purity-selected square region in image space, and its corresponding feature embedding is treated as a structured visual token in downstream models. On the one hand, parallel training and parallel inference typically require a consistent number of tokens across samples; on the other hand, ViT-style models naturally tokenize images on regular grids. Motivated by this, we instantiate the general Granular Ball Computing framework (Eq.~\eqref{equ:GBC}) with a \emph{non-standard granular ball}: instead of restricting balls to spheres, we represent each granular unit by an axis-aligned square region, denoted as $GB_i=\bigl(X_i,h_i\bigr)$, where $X_i\subset X$ is the set of pixels covered by the region and $h_i$ is its side length. This square geometry seamlessly tiles the image plane and guarantees strict cross-scale alignment, which facilitates hierarchical inheritance and mask propagation. Building on this formulation, we cast multi-granularity granular-ball token generation as a parallelizable selection process under a quality constraint (Eq.~\eqref{equ:GBflip}), and adopt a coarse-to-fine hierarchical selection and refinement strategy to produce structured superpixel tokens. Finally, cross-scale masks guide region feature extraction and enable consistent cross-scale alignment and feature fusion, yielding a multi-granularity visual representation that can be directly used by downstream models.
\subsection{Granular Ball Superpixel Tokenization in Image Space}
\label{sec:gb_tokens}
GBC provides a general paradigm that replaces point-wise learning on \(X\) with learning on a set of granular units \(GB=\{GB_i\}_{i=1}^{m}\), where each granular ball can be written as \(GB_i=(X_i,c_i,\vec{\theta}_i)\) and is not restricted to standard spheres (e.g., rectangles/ellipsoids). Its general form is
\begin{equation}\label{equ:GBC}
\begin{aligned}
&\quad \quad \quad \quad f(X,\vec{\alpha}) \longrightarrow  g(GB,\vec{\beta})\\
&\min_{c_i,\vec{\theta_{i}},\vec{\beta},m} \ \mathcal{J}\big(-Cov(GB),-Comp(GB)\big) + m \\
& \text{s.t.} \ \ quality(GB_i) \geq \phi(X),\ i = 1,2,\ldots,m, \\
& \qquad\ \ \ \ C(GB,X)\geq  0.
\end{aligned}
\end{equation}

In this work, we instantiate Eq.~(\ref{equ:GBC}) for superpixel generation by constructing granular balls directly in the input image space \(X\) with a square geometry. In particular, we interpret each granular ball as a superpixel region, and the resulting multi-granularity granular balls form a structured superpixel decomposition of the image. To enable parallel computation, we require a fixed number of superpixel tokens per image; meanwhile, ViT-style pipelines typically operate on regular square patches. Therefore, we adopt a non-standard granular ball represented by an axis-aligned square region,
\begin{equation}
GB_i=\bigl(X_i,h_i\bigr),
\end{equation}
where \(X_i\subset X\) denotes the pixels covered by the \(i\)-th square region and \(h_i\) denotes its side length. Our square granular-ball superpixels seamlessly tile the whole image plane, yielding full coverage without gaps; their compactness is explicitly controlled by \(h_i\): smaller \(h_i\) produces finer superpixels with stronger locality, while larger \(h_i\) yields coarser superpixels that emphasize global summarization. This design further guarantees strict spatial alignment across scales, which is convenient for enforcing hierarchical inheritance.

Under this instantiation, the construction of multi-granularity superpixel tokens is formulated as a parallelizable generation problem:
\begin{equation}\label{equ:GBflip}
\begin{aligned}
& \min_{h_{i},\vec{\beta},m}\ \ m \\
& \text{s.t.}\ \ quality(GB_i)\ge T,\quad i=1,2,\ldots,m,
\end{aligned}
\end{equation}
where \(T\) is a threshold hyper-parameter and \(\vec{\beta}\) denotes learnable parameters in the downstream model.

Granular ball superpixel tokens are generated in a coarse-to-fine manner rather than a one-shot fixed grid. At a given scale, the image is partitioned into a block set \(\{GB_i\}_{i=1}^{s}\). For each block \(GB_i\), we compute the mean intensity \(m_{GB_i}\) of its central \(2\times2\) pixels and measure block purity (quality) as
\begin{equation}\label{eq:quality_img}
\mathrm{quality}(GB_i)=\frac{1}{|GB_i|}\sum_{p\in GB_i}\big\|x_p-m_{GB_i}\big\|.
\end{equation}

Blocks that satisfy the purity constraint are retained as superpixels and recorded in a Boolean mask, whereas the remaining regions are further subdivided at finer scales. This coarse-to-fine process preserves parent--child inheritance and maintains strict spatial alignment across scales. Finally, the cross-scale masks guide region-wise feature extraction and enable consistent cross-scale alignment and feature fusion, producing a structured multi-granularity superpixel token representation. From a superpixel perspective, high-purity regions are typically captured by larger, coarser superpixels, while complex or heterogeneous areas are progressively refined into smaller, finer superpixels.

\subsection{Granular Ball Superpixel Generation}
\label{sec:gb_superpixel}

In this work, we reformulate superpixel generation from the perspective of GBC, where superpixels are treated as multi-granularity granular units constructed directly in the input image space \(X\). The objective is to obtain a structured decomposition that is interpretable and hierarchically organized, while remaining compatible with GPU-parallel processing. Unlike conventional superpixel methods that produce irregular regions, we represent each superpixel using an axis-aligned square region. This design guarantees strict spatial alignment across scales, avoids sample-dependent irregular shapes, and enables fixed-cardinality tokenization that can be seamlessly integrated into transformer-style pipelines.

At scale \(\ell\in\{1,\ldots,L\}\), the image is partitioned into a regular grid of resolution \(r_\ell\times r_\ell\), producing a candidate set \(\{GB_i^{(\ell)}\}_{i=1}^{r_\ell^2}\). Each candidate superpixel is represented by an axis-aligned square region and defined as
\begin{equation}
GB_i^{(\ell)}=\bigl(X_i^{(\ell)}, s_\ell\bigr),
\end{equation}
where \(X_i^{(\ell)}\subset X\) denotes the pixels covered by the \(i\)-th square at scale \(\ell\), and \(s_\ell\) is the side length. By construction, all candidates at a given scale tile the image plane without gaps, and the grids at different scales are naturally aligned, which provides a convenient basis for enforcing parent--child inheritance.

To evaluate the homogeneity of each candidate superpixel, we compute a robust center statistic from a small central window \(\mathcal{C}(GB_i^{(\ell)})\) (e.g., a \(k\times k\) window) as
\begin{equation}
m_i^{(\ell)}=\frac{1}{|\mathcal{C}(GB_i^{(\ell)})|}\sum_{p\in \mathcal{C}(GB_i^{(\ell)})}x_p,
\end{equation}
where \(x_p\in\mathbb{R}^3\) is the RGB intensity at pixel \(p\). Based on \(m_i^{(\ell)}\), we define a purity score as a thresholded within-block consistency measure:
\begin{equation}
\label{eq:purity_sp}
q^{(\ell)}_i
= \frac{1}{|GB_i^{(\ell)}|}\sum_{p\in GB_i^{(\ell)}}
\mathbb{I}\!\left(\left\|x_p - m_i^{(\ell)}\right\|_{1} < \tau\right),
\end{equation}
where \(\|\cdot\|_1\) denotes the \(\ell_1\) norm over RGB channels and \(\tau\) is a threshold controlling the tolerance to intra-superpixel variations. A larger \(q_i^{(\ell)}\) indicates that more pixels within the region are consistent with its center statistic, implying a more homogeneous superpixel.

Granular-Ball superpixels are generated in a coarse-to-fine manner. At coarse scales, homogeneous regions (with high purity) are directly retained as larger superpixels, whereas non-homogeneous regions are progressively subdivided into smaller candidates at finer scales. Concretely, a subset of high-purity candidates is selected at each coarse scale to represent regions that are already sufficiently consistent. The selected regions are then expanded to the next finer grid to mark them as covered, and the subsequent selection at finer scales is restricted to uncovered positions. Formally, we define an expansion operator \(\mathcal{E}(\cdot)\) that maps a selection/coverage mask on the \(\ell\)-th grid to a covered mask on the \((\ell+1)\)-th grid by assigning each selected square to its spatially aligned child squares, thereby enforcing parent--child inheritance and exact cross-scale alignment. After processing all scales, the remaining uncovered regions at the finest grid are retained to complete the tiling, ensuring full image coverage.

The above hierarchical refinement yields a structured multi-granularity superpixel set with a fixed total number of regions, determined by the selection budgets at coarse scales and the completion rule at the finest scale. Consequently, the resulting Granular-Ball superpixel decomposition simultaneously satisfies seamless coverage of the image plane, cross-scale alignment and inheritance, purity-driven adaptivity, and fixed-cardinality representation for downstream models. The overall procedure is summarized in Algorithm~\ref{alg:gb_superpixel}.

\begin{algorithm}[t]
\caption{Granular Ball Superpixel Generation (coarse-to-fine)}
\label{alg:gb_superpixel}
\begin{algorithmic}[1]
\Require Image \(X\), number of stages \(L\), grid sizes \(\{r_\ell\}_{\ell=1}^{L}\), side lengths \(\{s_\ell\}_{\ell=1}^{L}\), center window size \(k\), threshold \(\tau\), selection budgets \(\{K_\ell\}_{\ell=1}^{L-1}\)
\Ensure Multi-granularity superpixel set \(GB=\bigcup_{\ell=1}^{L}\{GB_i^{(\ell)}\}\) with fixed cardinality
\State Initialize covered mask on the coarsest grid: \(\tilde{M}^{(0)} \leftarrow \mathbf{0}\)
\For{\(\ell = 1\) \textbf{to} \(L\)}
    \State Partition \(X\) into \(r_\ell \times r_\ell\) square candidates \(\{GB_i^{(\ell)}\}_{i=1}^{r_\ell^2}\)
    \For{each candidate \(GB_i^{(\ell)}\)}
        \State Compute center statistic \(m_i^{(\ell)}\) using a \(k\times k\) window \(\mathcal{C}(GB_i^{(\ell)})\)
        \State Compute purity \(q_i^{(\ell)}\) by Eq.~\eqref{eq:purity_sp}
    \EndFor
    \If{\(\ell < L\)}
        \State Select \(M^{(\ell)}\) from uncovered positions \(\neg\tilde{M}^{(\ell-1)}\) using a budgeted policy (e.g., Top-\(K_\ell\) on \(q^{(\ell)}\))
        \State Update coverage for the next stage: \(\tilde{M}^{(\ell)} \leftarrow \mathcal{E}\big(\tilde{M}^{(\ell-1)} \lor M^{(\ell)}\big)\)
    \Else
        \State Set \(M^{(L)} \leftarrow \neg\tilde{M}^{(L-1)}\) \Comment{retain remaining finest candidates to complete tiling}
    \EndIf
\EndFor
\State Output selected superpixels \(GB \leftarrow \{GB_i^{(\ell)} \mid M^{(\ell)}_i = 1,\ \ell=1,\ldots,L\}\)
\end{algorithmic}
\end{algorithm}

\subsection{Time Complexity Analysis}
\label{sec:complexity}

We analyze the computational complexity of the proposed Granular-Ball superpixel generation in terms of the image size and the number of scales. Let the input image contain \(N=H\cdot W\) pixels and let the method use \(L\) grid scales \(\{r_\ell\times r_\ell\}_{\ell=1}^{L}\). At scale \(\ell\), the image is partitioned into \(r_\ell^2\) square candidates, and each candidate contains approximately \(N/r_\ell^2\) pixels.

The purity computation in Eq.~\eqref{eq:purity_sp} requires (i) computing the center statistic from a constant-size window \(\mathcal{C}(\cdot)\), which costs \(O(1)\) per candidate, and (ii) evaluating the thresholded consistency for all pixels within the candidate. Summing over all candidates at the same scale, every pixel is visited a constant number of times, yielding a per-scale cost of \(O(N)\). Therefore, the total cost of purity evaluation across all scales is
\begin{equation}
\sum_{\ell=1}^{L} O(N) = O(LN).
\end{equation}

The hierarchical selection and mask update mainly involve ranking or selecting candidates based on \(\{q_i^{(\ell)}\}\) and expanding the selected mask to the next scale. If a full sort is used, selection at scale \(\ell\) costs \(O(r_\ell^2 \log r_\ell^2)\); alternatively, a Top-\(K\) selection can be implemented in \(O(r_\ell^2)\) time. The mask expansion \(\mathcal{E}(\cdot)\) is linear in the number of grid cells and thus costs \(O(r_{\ell+1}^2)\). These terms are typically dominated by the \(O(N)\) pixel-wise purity evaluation when \(N\) is large, but for completeness we can write
\begin{equation}
T_{\text{total}}
= O(LN)\;+\;\sum_{\ell=1}^{L-1} O(r_\ell^2 \log r_\ell^2)\;+\;\sum_{\ell=1}^{L-1} O(r_{\ell+1}^2),
\end{equation}
or \(O(LN)+\sum_{\ell=1}^{L-1}O(r_\ell^2)\) when using linear-time Top-\(K\) selection.

In practice, \(L\) is small (a few stages), and the proposed method scales linearly with the number of pixels \(N\), making it efficient for high-resolution images and suitable for GPU-parallel execution.

% ---------- Explanation Paragraph (can be placed right after Table~\ref{tab:superpixel_complexity}) ----------
\noindent\textbf{Discussion.}
As shown in Table~\ref{tab:superpixel_complexity}, many superpixel methods scale approximately linearly with the number of pixels $N$, yet their computational patterns differ substantially. Graph partitioning and energy-optimization methods (e.g., Ncuts and ERS) rely on global graph optimization or spectral solvers and thus typically incur superlinear cost at high resolutions. Local clustering and region-evolution methods (e.g., SLIC, SEEDS, and TurboPixels) are near-linear but often require iterative pixel-level updates, yielding a complexity that grows with the iteration number $I$. End-to-end trainable approaches (e.g., SSN and SpixelFCN) also scale linearly with $N$, but their main overhead is dominated by network forward inference as well as the operations for assigning pixels to superpixels and aggregating features over the resulting regions.

% ---------- Complexity Table (place in Method section, e.g., after Time Complexity Analysis) ----------
\begin{table}[t]
\centering
\caption{Asymptotic time complexity comparison of representative superpixel methods. 
$N{=}HW$ denotes the number of pixels, $I$ is the number of iterations, and $L$ is the number of coarse-to-fine stages.}
\label{tab:superpixel_complexity}
\setlength{\tabcolsep}{5pt}
\renewcommand{\arraystretch}{1.05}
\scriptsize
\resizebox{0.95\linewidth}{!}{%
\begin{tabular}{l c}
\toprule
\textbf{Method} & \textbf{Time Complexity} \\
\midrule
Ncuts~\cite{shi2000normalized} & Superlinear in $N$ (global spectral/graph optimization) \\
ERS~\cite{liu2011entropy} & Typically superlinear or $O(N\log N)$ (solver-dependent) \\
SLIC & $O(NI)$ \\
SEEDS~\cite{van2012seeds} & $O(NI)$ (near-linear with iterative updates) \\
Watershed~\cite{vincent1991watersheds} & $O(N)$ \\
TurboPixels~\cite{levinshtein2009turbopixels} & $O(NI)$ \\
SSN~\cite{jampani2018superpixel} & $\mathrm{CNN}(N)+O(NI)$ \\
SpixelFCN~\cite{yang2020superpixel} & $\mathrm{FCN}(N)+O(N)$ \\
\textbf{Ours} & $O(LN)$\\
\bottomrule
\end{tabular}%
}
\end{table}

In contrast, our method has a worst-case complexity of $O(LN)$, where $L$ is a small number of coarse-to-fine stages (typically $L{=}3$), and in practice the cost is further reduced by refining only uncovered regions.
More importantly, our output is a fixed-budget, grid-aligned set of square tokens together with cross-scale masks, which can be plugged into ViT/RT-DETR style pipelines without additional irregular-region adaptation.
Therefore, the advantage of our method lies not in reducing the asymptotic order below linear, but in providing a structured, budget-controllable, and GPU-parallel friendly tokenization that yields a better accuracy--efficiency trade-off in downstream models.

\section{Experiments} 
To validate the effectiveness and generalization ability of the proposed method, we conduct systematic experiments on three representative vision tasks, including graph-based image classification, image--text retrieval, and object detection. We integrate the module into network architectures of different paradigms and perform both comparative evaluations and ablation studies on multiple public benchmarks. The results demonstrate consistent gains across tasks and architectures; moreover, under token pruning settings, our method achieves a more favorable accuracy--efficiency trade-off, highlighting its practical value as a plug-and-play acceleration and structured representation module.

\subsection{Image Classification Task}
\textbf{Datasets and Metrics.} We evaluate the proposed method on two standard image classification benchmarks, \textit{MNIST} and \textit{CIFAR-10}. MNIST consists of single-channel grayscale handwritten digit images, whereas CIFAR-10 contains three-channel color natural images. We report \textbf{Top-1 test accuracy} as the primary metric for all experiments. For qualitative inspection, Fig.~\ref{fig} compares our adaptive block partitioning with SLIC on both datasets, where our method tends to produce finer partitions around structurally complex regions while keeping coarser blocks in relatively homogeneous areas.
\begin{figure}[htbp]
\centering
\includegraphics[width=0.45\textwidth,keepaspectratio]{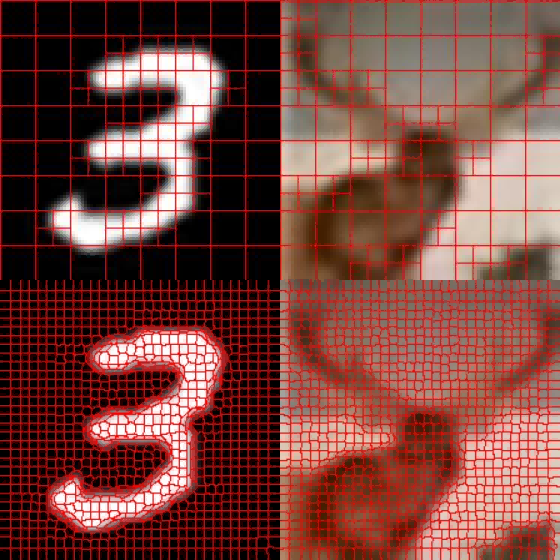} 
\caption{Our Method vs. SLIC on MNIST and CIFAR-10. Top: Our Method; Bottom: SLIC Method.}
\label{fig}
%\vspace{-1em}
\end{figure}

\textbf{Experimental Setup.} 
\textit{MNIST.} Since MNIST has only one input channel, we replace the ResNet feature extractor with lightweight convolutional layers to avoid introducing unnecessary capacity differences, and feed the extracted features into the ViG module~\cite{han2022vision} for classification. We adopt the Tiny configuration of ViG with the number of blocks set to 1, while keeping other settings unchanged. The numbers of patch blocks at the coarse and fine granularities are set to 25 and 96, respectively. We use the Adam optimizer with a batch size of 64, and train the model for 1000 epochs with an initial learning rate of 0.001.

\textit{CIFAR-10.} For CIFAR-10, we employ a hybrid architecture combining ResNet and ViG modules: the ResNet branch uses the BasicBlock structure, while the ViG module keeps the same configuration as in the MNIST experiments. To accommodate the higher structural complexity of CIFAR-10 images, we set the numbers of patch blocks at two granularity levels to 45 and 76. Moreover, to examine the effect of model capacity, we construct three model scales: \textbf{Small} (ResNet14 + ViG1), \textbf{Base} (ResNet20 + ViG1), and \textbf{Large} (ResNet32 + ViG2). All models are trained using SGD with a batch size of 128 for 200 epochs. We adopt a cosine annealing scheduler with an initial learning rate of 0.1.

\textbf{Comparative Experiments and Analysis.} We compare against a diverse set of representative graph neural networks and graph Transformers, including Vanilla GCN~\cite{kipf2016semi}, GraphSAGE~\cite{hamilton2017inductive}, MoNet~\cite{monti2017geometric}, GatedGCN~\cite{bresson2017residual}, GAT~\cite{velivckovic2017graph}, GIN~\cite{xu2018powerful}, RingGNN~\cite{chen2019equivalence}, 3WLGNN~\cite{maron2019provably}, GraphCON~\cite{rusch2022graph}, GPS~\cite{rampavsek2022recipe}, EGT~\cite{hussain2022global}, GatedGCN + SSFG~\cite{zhang2022ssfg}, Cy2C-GNNs~\cite{choi2022cycle}, ARGNP~\cite{cai2022automatic}, ShapeGNN~\cite{cosma2023geometric}, EdgeGCN~\cite{zhang2023learning}, Exphormer~\cite{shirzad2023exphormer}, GRIT~\cite{ma2023graph}, NG~\cite{kofinas2024graph}, GRASP-GCN~\cite{casarin2024grasp}, GeoMix~\cite{zhao2024geomix}, TIGT~\cite{choi2024topology}, and GRI\_GAT~\cite{10996538}. Quantitative results are summarized in Table~\ref{tab:performance} (The best results are \textbf{boldfaced}.).

\begin{table}[!htbp]
\caption{Performance Comparison on MNIST and CIFAR10. The best results are \textbf{boldfaced}.}
\begin{center}
\scalebox{1}{
\begin{tabular}{cccc}
\toprule
\multirow{2}{*}{\vspace{-1em}\textbf{Model}} & \multirow{2}{*}{\vspace{-1em}\textbf{Date}} & \multicolumn{1}{c}{\textbf{MNIST}} & \multicolumn{1}{c}{\textbf{CIFAR10}} \\ 
\cmidrule(lr){3-3} \cmidrule(lr){4-4} & & \textbf{Test ACC} & \textbf{Test ACC} \\ 
\midrule
Vanilla GCN            & ICLR 2017    & 90.70  & 55.71   \\ 
GraphSage              &  NIPS 2017  & 97.31   & 65.76   \\ 
MoNet                  &  CVPR 2017  & 90.80   & 54.65   \\ 
GatedGCN               &  Arxiv   & 97.34    & 67.31   \\ 
GAT                    &  ICLR 2018  & 95.53    & 64.22   \\ 
GIN                    &  ICLR 2019  & 96.48    & 55.25   \\ 
RingGNN                & NeurIPS 2019   & 91.86   & 39.16   \\ 
3WLGNN                 &  NeurIPS 2019  & 95.07    & 59.17   \\ 
GraphCON-GCN           &  ICML 2022  & 98.68       & -        \\ 
GraphCON-GAT           &  ICML 2022  & 98.91      & -        \\ 
GPS               &  NeurIPS 2022  & 98.05   & 72.29   \\
EGT                    & KDD 2022   & 98.17    & 68.70  \\
GatedGCN + SSFG        &  TNNLS 2022  &  97.99       & 71.94   \\
% ResNet20               & CVPR2016   & -        & 91.25   \\ 
Cy2C-GNNs              &  ICLR 2022 & 97.77         &  64.29   \\ 
ARGNP                  & CVPR 2022   & -       &   73.9   \\ 
ShapeGNN               & ICCV 2023   & -      & 80.40  \\
EdgeGCN                & TNNLS S2023 &  98.43        & 76.13   \\
Exphormer              & ICML 2023   &  98.55       & 74.75   \\
GRIT                   &  ICML 2023  &  98.11     & 76.47   \\
NG-GNN                 & ICLR 2024 &  94.7       & 64.37   \\
NG-T                   & ICLR 2024  & 97.3        & 60.79   \\
GRASP-GCN              &  CVPR 2024  & -        & 87.9   \\
GeoMix                 & SIGKDD 2024 &  -       & 71.3   \\
TIGT                   &  arXiv  &  98.23       & 73.96   \\
GRI\_GAT                &  TIP 2025 & 98.82      & 76.04    \\
\multirow{3}{*}{\textbf{Ours}} & \multirow{3}{*}{-} & \multirow{3}{*}{\textbf{99.40}}  & \text{91.45} (S) \\ 
& & & 92.40 (B) \\
& & & \textbf{93.47} (L) \\
\bottomrule
\end{tabular}
}
%\vspace{-1em}
\label{tab:performance}
\end{center}
\end{table}

Table~\ref{tab:performance} reports the test accuracies on MNIST and CIFAR-10. Overall, our method achieves a test accuracy of 99.40\% on MNIST. On CIFAR-10, it obtains 91.45\%, 92.40\%, and 93.47\% under the Small, Base, and Large configurations, respectively. These results indicate that the proposed approach yields consistent gains across datasets with distinct characteristics, and that increasing model capacity leads to improved classification accuracy.

From a methodological perspective, graph-based models are advantageous for modeling local relationships, yet their performance often depends on graph construction and relatively shallow feature extraction, making them sensitive to input resolution and structural cues. For instance, ShapeGNN incorporates contour information, but the low resolution of CIFAR-10 limits the availability of explicit shape cues, which can restrict its performance gains. In contrast, our method introduces a granular-ball mechanism to organize images into adaptive square blocks, leveraging a purity criterion and multi-granularity partitioning to dynamically allocate representation granularity across homogeneous and structurally complex regions. This design helps suppress redundant background/noisy regions while preserving more discriminative local structures, thereby improving the learned representations for classification.

\subsection{Image-Text Retrieval Task}

\textbf{Task and Model.}
We evaluate the proposed method on image--text retrieval, including both image-to-text and text-to-image directions, to assess its cross-modal alignment capability.
All experiments are built upon the FLIP framework \cite{dai202415m}.
FLIP follows the dual-tower design of CLIP and is developed as a domain-specific vision--language pre-training model for the \emph{human} domain, aiming to achieve cross-domain alignment on general image--text instruction data.
Its image encoder consists of a \emph{granular-ball square-superpixel module} and a ViT backbone, while the text encoder is a standard BERT model.
Based on this framework, we integrate our granular-ball square-superpixel module into the image-side encoder to enable multi-granularity semantic alignment in the feature space.
For controlled evaluation, we also construct an ablated variant by removing the proposed module, denoted as FLIP*.

\textbf{Datasets and Metrics.}
We conduct evaluations on two datasets, CelebA~\cite{liu2018large} and MM-CelebA~\cite{xia2021tedigan}.
CelebA is a relatively standard face-attribute vision--language dataset, whereas MM-CelebA poses more challenging cross-modal and cross-distribution variations.
We adopt the standard Recall@K metrics for retrieval, i.e., R@1, R@5, and R@10, and report results for both image-to-text and text-to-image retrieval.

\textbf{Experimental Setup.}
To follow the FLIP pre-training protocol and ensure fair comparisons, we train a new model from scratch under a unified training recipe with our module integrated (denoted as \textbf{Ours}).
Concretely, the convolutional branch in our module uses a frozen, pre-trained ResNet-50 to extract basic local features, and we set the patch-block numbers to 24, 50, and 200 at different granularities to construct granular-ball square superpixels.
To isolate the gains brought by the proposed module and mitigate potential influence from the convolutional backbone, we additionally train a counterpart model without the granular-ball square-superpixel module, denoted as \textbf{FLIP*}.
Moreover, to avoid unfair comparisons caused by different training pipelines, we define \textbf{FLIP} as the official released model (or the results reported in the original paper and re-evaluated under our evaluation pipeline) and treat it as a \emph{reference}.
In contrast, \textbf{FLIP*} and \textbf{Ours} form a controlled pair that is re-trained under the same implementation and recipe.
In particular, FLIP* and Ours share identical settings in data processing, optimizer, learning-rate schedule, training epochs, batch size, and the visual backbone (uniformly using ViT-Base/16), such that their difference can strictly quantify the contribution of our module.
We note that minor metric fluctuations between the official FLIP and our re-trained FLIP* may occur due to differences in implementation details and data preprocessing, which is expected.
For downstream evaluation, we follow FLIP's standard two-stage protocol: (i) zero-shot retrieval evaluation; and (ii) lightweight adaptation for retrieval comparison, where in Table~4 we freeze the backbone and only train a linear layer.

\textbf{Comparative Experiments and Analysis.}
\emph{(1) Zero-shot comparison.}
Table~\ref{tab:base} reports the zero-shot retrieval results of FLIP, FLIP*, and Ours on CelebA and MM-CelebA.
Overall, our method consistently outperforms both FLIP and FLIP* on CelebA for both retrieval directions, indicating that incorporating the multi-granularity granular-ball structure improves cross-modal alignment even without task-specific fine-tuning.
On the more challenging MM-CelebA benchmark, although FLIP* may exhibit minor fluctuations on certain metrics, our method achieves the best overall or more robust performance, suggesting strong cross-dataset generalization brought by the proposed module.

\begin{table}[htbp]
\centering
\caption{Performance Comparison on CelebA and MM-CelebA Zero-Shot Tasks. The best results are \textbf{boldfaced}.}
\setlength{\tabcolsep}{3pt} % 减小列间距
\scalebox{1}{
\begin{tabular}{cccccccc}
\toprule 
\multirow{2}{*}{\vspace{-1em}\textbf{Dataset}} & \multirow{2}{*}{\vspace{-1em}\textbf{Model}} & \multicolumn{3}{c}{\textbf{Image=\scalebox{0.9}{$>$}Text}} & \multicolumn{3}{c}{\textbf{Text=\scalebox{0.9}{$>$}Image}} \\
\cmidrule(lr){3-5}\cmidrule(lr){6-8}
& & \textbf{R@1} & \textbf{R@5} & \textbf{R@10} & \textbf{R@1} & \textbf{R@5} & \textbf{R@10} \\ 
\midrule
\multirow{4}{*}[0.15cm]{CelebA} 
& FLIP & 21.99 & 43.9 & 57.63 & 21.84 & 45.64 & 56.34  \\ 
& FLIP* & 22.64 & 43.96 & 57.77 & 22.42 & 46.12 & 56.48  \\ 
& \textbf{Ours}   & \textbf{23.08} & \textbf{45.58} & \textbf{58.42} & \textbf{22.66}  & \textbf{47.98}  & \textbf{58.54}       \\  
\midrule
\multirow{4}{*}[0.15cm]{MM-CelebA} 
& FLIP  & 7.63  & 19.15 & 26.06  & 8.85 & 21.5 & 28.86 \\ 
& FLIP* & 8.41  & 21.75 & 29.18  & 9.2 & 23.3 & 31.58          \\ 
& \textbf{Ours}   & \textbf{9.01} & \textbf{22.43} & \textbf{30.31} & \textbf{9.46}  & \textbf{23.43} & \textbf{31.68}         \\ 
\bottomrule
\end{tabular}
}
\label{tab:base}
\end{table}

\emph{(2) lightweight adaptation.}
Table~\ref{tab:linear} further compares our method with ALIGN \cite{li2021align}, BLIP \cite{li2022blip}, CLIP \cite{radford2021learning}, and FLIP \cite{dai202415m} under a protocol that freezes the backbone and fine-tunes only a linear layer, so as to evaluate the transferability of pre-trained representations.
Our method achieves the best or near-best retrieval performance on both CelebA and MM-CelebA.
In particular, on MM-CelebA we outperform the FLIP baseline in both image-to-text and text-to-image retrieval, demonstrating that the proposed module can better leverage fine-grained regional features and enhance structural representation in images, thereby improving overall cross-modal retrieval performance.

\begin{table}[htbp]
\centering
\caption{Performance Comparison on CelebA and MM-CelebA. All pretrained model backbones are frozen, with only the linear layer being fine-tuned. The best results are \textbf{boldfaced}.}
\setlength{\tabcolsep}{3pt} % 减小列间距
\scalebox{1}{
\begin{tabular}{cccccc}
\toprule 
\multirow{2}{*}{\vspace{-1em}\textbf{Dataset}} & \multirow{2}{*}{\vspace{-1em}\textbf{Model}} & \multicolumn{2}{c}{\textbf{Image=\scalebox{1}{$>$}Text}} & \multicolumn{2}{c}{\textbf{Text=\scalebox{1}{$>$}Image}} \\
\cmidrule(lr){3-4}\cmidrule(lr){5-6}
& & \textbf{R@5} & \textbf{R@10}& \textbf{R@5} & \textbf{R@10} \\ 
\midrule
\multirow{4}{*}[-0.2cm]{\centering CelebA} 
& ALIGN  & 6.99  & 11.09  & 5.08  & 8.65  \\ 
& BLIP  &24.56  &34.09   & 24.35 &34.15   \\
& CLIP   &21.42  &30.3  &26.52  &37.22   \\
& FLIP   & 50.04  & 62.04  & \textbf{50.96}  & \textbf{62.98}  \\
& \textbf{Ours} & \textbf{50.12}  & \textbf{62.36} & 50.78  & 62.2 \\  
\midrule
\multirow{4}{*}[-0.2cm]{\centering MM-CelebA} 
& ALIGN  & 8.1  & 12.61 & 6.6  & 10.53  \\ 
& BLIP   &21.56  &31.25  & 23.83 & 33.38  \\
& CLIP   &13.63  &21.06  &14.41  &22.48   \\
& FLIP  & 26.1  & 35.75  & 27.05  & 37.23  \\
& \textbf{Ours}  & \textbf{27.4}  & \textbf{37.5} & \textbf{27.93}  & \textbf{37.56} \\ 
\bottomrule
\end{tabular}
}
\vspace{-0.17em}
\label{tab:linear}
\end{table}

\subsection{Object Detection Task}                                                                     
\textbf{Task and Model.} We evaluate our granular-ball square-superpixel token pruning strategy on object detection by integrating it into RT-DETR (Real-Time DEtection TRansformer)~\cite{lv2023detrs}. RT-DETR is a representative real-time end-to-end detector with a hybrid encoder that combines CNN and Transformer modules, providing a suitable testbed for analyzing the accuracy--efficiency trade-off of encoder-side token pruning.

\begin{figure}[htbp]                                                                       
  \centering                                                                                                                         
  \includegraphics[width=0.48\textwidth,keepaspectratio]{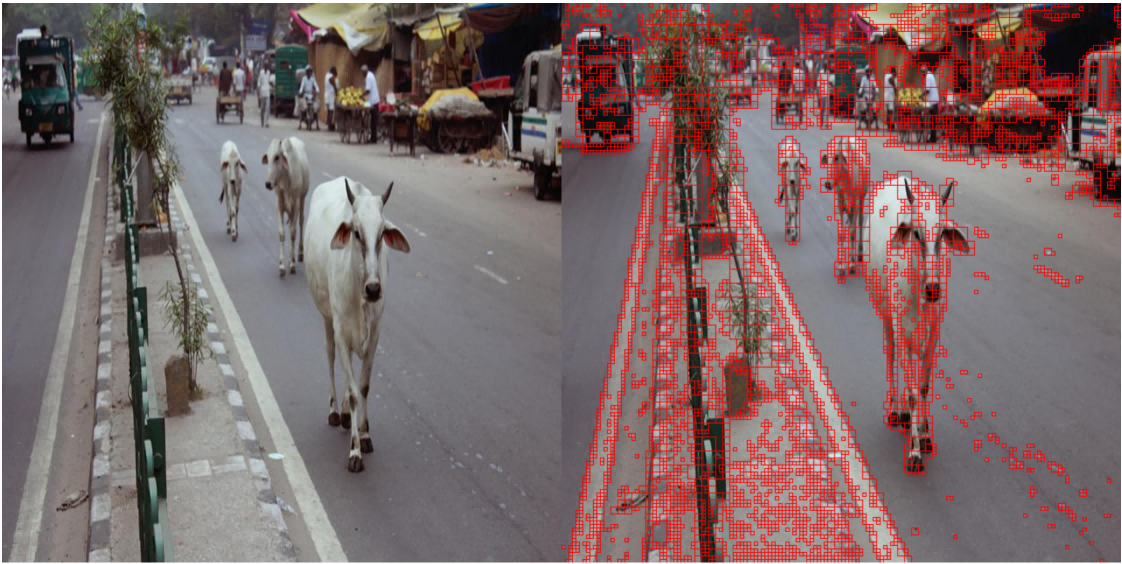}                                                                
  \caption{Visualization of multi-granularity token selection on COCO image. Left: Original image with objects at various scales (cows and pedestrians).    
  Right: Selected tokens represented by square blocks of varying sizes. Our method adaptively assigns larger blocks to homogeneous background regions and   
  smaller blocks to object-dense areas, effectively reducing tokens while preserving critical information for detection.}                                   
  \label{fig:detection_multigrain}                                                                                                                          
  \end{figure}

\textbf{Dataset and Metrics.} We conduct experiments on the MS COCO 2017 dataset~\cite{lin2014microsoft}, which contains 118K training images and 5K validation images across 80 object categories. We report standard COCO detection metrics on the validation set, including AP, AP$_{50}$, and AP$_{75}$, as well as scale-specific AP for small, medium, and large objects, denoted as AP$_S$, AP$_M$, and AP$_L$, respectively.

\textbf{Experimental Setup.} We follow the standard RT-DETR-R50 configuration, using ResNet-50 with deformable convolutions (ResNet50-vd) as the backbone. The model is trained with a 6$\times$ schedule (72 epochs) using the AdamW optimizer, with an initial learning rate of $1\times 10^{-4}$ and a batch size of 8. Except for the token pruning strategy, all other training settings are kept identical to ensure that the results reflect the effect of the pruning module itself.

\textbf{Token Pruning in RT-DETR.} We apply our pruning module to the hybrid encoder of RT-DETR. With an input resolution of $640\times640$, the original feature map corresponds to a $20\times20$ grid, i.e., 400 tokens in total. We further study different pruning ratios by removing 50, 100, and 200 tokens, resulting in token retention ratios of 87.5\%, 75.0\%, and 50.0\%, respectively.

Specifically, we adopt a block-based multi-stage (coarse-to-fine) selection strategy with three stages using block sizes of [32, 16, 8], and progressively select high-purity regions according to the purity score defined in Eq.~\ref{eq:quality_img}. The purity threshold $\tau$ is set to 10. Fig.~\ref{fig:detection_multigrain} visualizes the selected tokens: our method tends to assign larger blocks to homogeneous background regions while allocating finer, smaller blocks to object-dense areas, thereby reducing the number of tokens while preserving critical information for detection.

\textbf{Comparative Experiments and Analysis.} To ensure a fair comparison, we adopt RT-DETR-R50 as our baseline and integrate our multi-granularity, purity-driven token selection module into its hybrid encoder under the same training configuration. We compare different token retention ratios: the baseline RT-DETR uses 400 tokens (a $20\times20$ grid) with a $640\times640$ input, while our pruning setting reduces the number of tokens to 200 (removing 200 tokens, i.e., 50\% compression).

Meanwhile, Table~\ref{tab:detection_coco_all} summarizes the performance of detectors from different categories on COCO, including real-time one-/two-stage detectors (e.g., the YOLO family such as YOLOv6 \cite{li2023yolov6}, YOLOv7 \cite{wang2023yolov7} and YOLOv8 \cite{jocher2023yolov8}) and end-to-end Transformer-based detectors (the DETR family, including DETR-DC5 \cite{carion2020end}, Anchor-DETR-DC5 \cite{wang2022anchor}, Conditional-DETR-DC5 \cite{meng2021conditional}, Efficient-DETR \cite{yao2021efficient}, SMCA-DETR \cite{gao2021fast}, Deformable-DETR \cite{zhu2020deformable}, DAB-Deformable-DETR \cite{liu2022dab}, DAB-Deformable-DETR++ \cite{liu2022dab},  DN-Deformable-DETR \cite{li2022dn}, DN-Deformable-DETR++ \cite{li2022dn}, DINO-Deformable-DETR \cite{zhang2022dino} and RT-DETR \cite{lv2023detrs}). Overall, our method is highly competitive within the end-to-end detection paradigm: even when retaining only 200 tokens (removing 200 out of the 400 tokens in RT-DETR, i.e., 50\% compression), we still achieve 52.3 AP, outperforming most end-to-end DETR variants in the table (e.g., DETR-DC5, Anchor-DETR, Conditional-DETR, Efficient-DETR, SMCA-DETR, and Deformable-DETR). These results indicate that the proposed granular-ball square-superpixel token selection effectively preserves the most critical regions for detection while reducing redundant background tokens, leading to a better accuracy--efficiency trade-off within an end-to-end framework.

  % 这个是新增的总的实验
% in IEEEtran two-column paper: use table* to span two columns
\begin{table*}[t]
\centering
\caption{Performance comparison on COCO object detection. The best results are \textbf{boldfaced}.}
\label{tab:detection_coco_all}
\setlength{\tabcolsep}{4pt}
\renewcommand{\arraystretch}{1.12}
\scriptsize
\resizebox{\textwidth}{!}{
\begin{tabular}{l c c c c c c c c c c}
\toprule
Model & Backbone & \#Epochs & \#Params (M) & GFLOPs &
$AP^{val}$ & $AP^{val}_{50}$ & $AP^{val}_{75}$ & $AP^{val}_{S}$ & $AP^{val}_{M}$ & $AP^{val}_{L}$ \\
\midrule
\multicolumn{11}{l}{\textit{Real-time Object Detectors}} \\
\midrule
YOLOv6-L  & -- & 300 & 59 & 150 & 52.8 & 70.3 & 57.7 & 34.4 & 58.1 & 70.1 \\
YOLOv7-L  & -- & 300 & 36 & \textbf{104} & 51.2 & 69.7 & 55.5 & 35.2 & 55.9 & 66.7 \\
YOLOv7-X  & -- & 300 & 71 & 189 & 52.9 & 71.1 & 57.4 & \textbf{36.9} & 57.7 & 68.6 \\
YOLOv8-L  & -- & --  & 43 & 165 & 52.9 & 69.8 & 57.5 & 35.3 & 58.3 & 69.8 \\
YOLOv8-X & -- & --  & 68 & 257 & 53.9 & 71.0 & \textbf{58.7} & 35.7 & \textbf{59.3} & \textbf{70.7} \\
\midrule
\multicolumn{11}{l}{\textit{End-to-end Object Detectors}} \\
\midrule
DETR-DC5  & R50  & 500 & 41 & 187 & 43.3 & 63.1 & 45.9 & 22.5 & 47.3 & 61.1 \\
% DETR-DC5 & R101 & 500 & 60 & 253 & 44.9 & 64.7 & 47.7 & 23.7 & 49.5 & 62.3 \\
Anchor-DETR-DC5  & R50  & 50  & 39 & 172 & 44.2 & 64.7 & 47.5 & 24.7 & 48.2 & 60.6 \\
% Anchor-DETR-DC5 & R101 & 50  & -- & --  & 45.1 & 65.7 & 48.8 & 25.8 & 49.4 & 61.6 \\
Conditional-DETR-DC5  & R50  & 108 & 44 & 195 & 45.1 & 65.4 & 48.5 & 25.3 & 49.0 & 62.2 \\
% Conditional-DETR-DC5  & R101 & 108 & 63 & 262 & 45.9 & 66.8 & 49.5 & 27.2 & 50.3 & 63.3 \\
Efficient-DETR  & R50  & 36  & \textbf{35} & 210 & 45.1 & 63.1 & 49.1 & 28.3 & 48.4 & 59.0 \\
% Efficient-DETR  & R101 & 36  & 54 & 289 & 45.7 & 64.1 & 49.5 & 28.2 & 49.1 & 60.2 \\
SMCA-DETR  & R50  & 108 & 40 & 152 & 45.6 & 65.5 & 49.1 & 25.9 & 49.3 & 62.6 \\
% SMCA-DETR  & R101 & 108 & 58 & 218 & 46.3 & 66.6 & 50.2 & 27.2 & 50.5 & 63.2 \\
Deformable-DETR  & R50 & 50 & 40 & 173 & 46.2 & 65.2 & 50.0 & 28.8 & 49.2 & 61.7 \\
DAB-Deformable-DETR & R50 & 50 & 48 & 195 & 46.9 & 66.0 & 50.8 & 30.1 & 50.4 & 62.5 \\
DAB-Deformable-DETR++ & R50 & 50 & 47 & --  & 48.7 & 67.2 & 53.0 & 31.4 & 51.6 & 63.9 \\
DN-Deformable-DETR  & R50 & 50 & 48 & 195 & 48.6 & 67.4 & 52.7 & 31.0 & 52.0 & 63.7 \\
DN-Deformable-DETR++ & R50 & 50 & 47 & --  & 49.5 & 67.6 & 53.8 & 31.3 & 52.6 & 65.4 \\
DINO-Deformable-DETR  & R50 & 36 & 47 & 279 & 50.9 & 69.0 & 55.3 & 34.6 & 54.1 & 64.6 \\
\midrule
\multicolumn{11}{l}{\textit{Real-time End-to-end Object Detector}} \\
\midrule
RT-DETR (400 Token) & R50  & 72 & 42 & 138 & \textbf{53.1} & \textbf{71.3} & 57.7 & 34.8 & 58.0 & 70.0 \\
Ours (200 Token) & R50 & 72 & 42 & 137.5 & 52.3 & 70.3 & 56.6 & 33.4 & 56.3 & 69.8 \\
\bottomrule
\end{tabular}
}
\end{table*}

In contrast, the YOLO series typically attains higher absolute AP in Table~\ref{tab:detection_coco_all}. This is mainly because its detection paradigm, training recipes, and extensive engineering optimizations (e.g., stronger feature pyramids, dense prediction heads, and various practical tricks) differ substantially from the end-to-end DETR line. Therefore, direct comparisons between the two families largely reflect paradigm differences rather than the superiority of a single module. Importantly, our work is positioned as a plug-and-play token pruning acceleration module within the RT-DETR framework: without materially altering the detector architecture, we halve the encoder tokens while keeping performance close to the RT-DETR baseline (53.1$\rightarrow$52.3), demonstrating practical value in reducing computational cost for real-time deployment.

Overall, the results in Table~\ref{tab:detection_coco_all} suggest that our method not only achieves significant token compression and potential computational savings on RT-DETR, but also remains strongly competitive among end-to-end detectors. Although there is still a gap compared with highly engineered real-time detectors such as YOLO, our approach can serve as a general efficiency enhancement strategy and can be further combined with stronger backbones or more advanced detection frameworks.

\textbf{Sensitivity to Pruning Ratio.} To further characterize how token pruning affects detection accuracy, we perform a sensitivity analysis under the same RT-DETR-R50 training setting with different token retention ratios. The results are reported in Table~\ref{tab:detection_coco}. Without pruning, RT-DETR achieves 53.1 AP, consistent with the original performance. When removing 50 tokens (12.5\% reduction), AP decreases to 52.5 (-0.6). Removing 100 tokens (25\% reduction) yields 52.3 AP (-0.8). Notably, when further removing 200 tokens (halving the tokens), AP remains 52.3, identical to the 100-token setting, exhibiting a clear ``performance plateau''. This indicates that once the most critical region tokens are retained, further discarding background tokens has a much smaller marginal impact on detection accuracy, demonstrating strong robustness and graceful degradation under aggressive pruning.

  \begin{table}[htbp]                                                                                                                                       
  \centering                                                                                                                                                
  \caption{Performance Comparison on COCO Object Detection with Different Token Pruning Ratios. All models use RT-DETR-R50 backbone trained on COCO 2017. The best results are \textbf{boldfaced}.}  
  \setlength{\tabcolsep}{3pt}                                                                                                                               
  \scalebox{1}{                                                                                                                                          
  \begin{tabular}{ccccccccc}                                                                                                                                 
  \toprule                                                                                                                                                  
  \textbf{Model} & \textbf{Tokens} & \textbf{Removed} & \textbf{AP} & \textbf{AP$_{50}$} & \textbf{AP$_{75}$} & \textbf{AP$_S$} & \textbf{AP$_M$} & \textbf{AP$_L$} \\                                                                                                                                        
  \midrule                                                                                                                                                  
  RT-DETR & 400 & 0 & \textbf{53.1} & \textbf{71.3} & \textbf{57.7} & 34.8 & \textbf{58.0} & \textbf{70.0} \\                                                                                            
  \textbf{Ours} & 350 & 50 & 52.5 & 70.6 & 56.5 & 34.2 & 56.6 & 69.8 \\                                                                                     
  \textbf{Ours} & 300 & 100 & 52.3 & 70.4 & 56.6 & \textbf{35.4} & 56.7 & 69.9 \\                                                                           
  \textbf{Ours} & 200 & 200 & 52.3 & 70.3 & 56.6 & 33.4 & 56.3 & 69.8 \\                                                                                    
  \bottomrule                                                                                                                                               
  \end{tabular}                                                                                                                                             
  }                                                                                                                                                         
  \label{tab:detection_coco}                                                                                                                                
  \end{table}                

\textbf{Impact Across Object Scales.} We observe different trends across object scales. Pruning affects large objects (AP$_L$) the least, with only a slight drop of about 0.1--0.2 across all pruning ratios, suggesting that our coarse-to-fine block selection can effectively cover the contiguous regions containing large objects and preserve the key tokens. In contrast, medium objects (AP$_M$) suffer a larger degradation (58.0$\rightarrow$56.3--56.7), likely because medium-sized targets require denser and more precise token coverage, and aggressive pruning partially sacrifices spatial granularity.

For small objects (AP$_S$), we observe a non-monotonic behavior: mild pruning (removing 50 tokens) slightly reduces AP$_S$ (34.8$\rightarrow$34.2), whereas moderate pruning (removing 100 tokens) improves AP$_S$ to 35.4 (+0.6 over the baseline), before dropping to 33.4 under more aggressive pruning (removing 200 tokens). This suggests that moderate pruning can suppress background clutter and noisy tokens, benefiting small-object recognition, while excessive pruning removes necessary local context and hurts small-object performance.

\textbf{Efficiency Analysis.}
The identical overall AP (52.3) obtained with 300 and 200 retained tokens in Table~\ref{tab:detection_coco} indicates that the encoder input sequence contains substantial redundancy. Once the proposed purity-driven selection mechanism has preserved the most informative and object-relevant tokens, many of the subsequently removed tokens mainly correspond to background regions or low-contribution areas, and therefore have only a limited effect on final detection accuracy. This observation suggests that the proposed multi-granularity structured token selection strategy can maintain strong performance stability even under relatively aggressive compression.

From the perspective of theoretical complexity, token pruning mainly reduces the self-attention computation in the encoder. Since the complexity of self-attention scales approximately quadratically with the number of tokens, reducing the token number from 400 to 300 changes the attention cost from \(O(400^2)\) to \(O(300^2)\), corresponding to an approximate 43.8\% reduction in encoder-side attention complexity. When the token number is further reduced to 200, the attention complexity becomes \(O(200^2)\), yielding a theoretical reduction of about 75.0\% relative to the baseline.

It should be noted, however, that these reductions mainly refer to the self-attention computation within the encoder, rather than the end-to-end FLOPs of the entire detector. In RT-DETR, the backbone, convolutional feature extraction, feature fusion, and decoder modules still account for a considerable portion of the total computation. As a result, even when the encoder token number is significantly reduced, the decrease in overall GFLOPs may remain relatively limited. Therefore, the main advantage of our method lies in effectively alleviating the attention burden caused by redundant encoder tokens, while preserving detection accuracy with only a small performance drop. This demonstrates a favorable accuracy--efficiency trade-off and highlights the practical potential of the proposed method for real-time end-to-end detection in resource-constrained scenarios.

\section{CONCLUSION}
This paper proposes an efficient superpixel-based plug-and-play module with a simple structure, ease of implementation, and strong portability, which can be conveniently integrated into existing deep learning frameworks. Without altering the overall backbone architecture, the module is, in principle, compatible with a broad range of Transformer models. By introducing multi-scale structural priors, it aims to improve the efficiency of feature representation and, to some extent, enhance model generalization. Specifically, we first partition the input image into multi-scale patches and compute the purity of each patch; then, a fixed number of patches are selected according to their purity to generate a mask matrix for interpolation. Next, the image is normalized, and a convolutional network is employed to extract features corresponding to patches at different scales. Finally, features from each stage are fused with their respective masks to form a multi-scale feature representation of the image. In the future, we plan to extend this method to more ViT-based vision tasks and systematically evaluate the effectiveness and applicability of superpixel structural priors across different tasks and architectures.

\bibliography{ref}
\bibliographystyle{IEEEtran}
% \bibliographystyle{named}
% \bibliography{ijcai25}

\end{document}